\pdfoutput=1
\documentclass{sigchi}

\usepackage[final]{pdfpages}
\usepackage{balance}  
\usepackage{graphics} 
\usepackage{times}    
\usepackage{url}      
\usepackage{graphicx}
\usepackage[T1]{fontenc}   
\usepackage{txfonts}
\usepackage{mathptmx}
\usepackage[pdflang={en-US},pdftex]{hyperref}
\usepackage{color}
\usepackage{booktabs}
\usepackage{textcomp}
\usepackage[hang,tight,footnotesize,sf,SF,small]{subfigure}
\usepackage{enumitem}
\usepackage{comment}

\makeatletter
\def\url@leostyle{%
  \@ifundefined{selectfont}{\def\UrlFont{\sf}}{\def\UrlFont{\small\bf\ttfamily}}}
\makeatother
\def\pprw{8.5in}
\def\pprh{11in}

\usepackage[pdftex]{hyperref}
\hypersetup{
pdftitle={SIGCHI Conference Proceedings Format},
pdfauthor={LaTeX},
pdfkeywords={SIGCHI, proceedings, archival format},
bookmarksnumbered,
pdfstartview={FitH},
colorlinks,
citecolor=black,
filecolor=black,
linkcolor=black,
urlcolor=black,
breaklinks=true,
}

\begin{document}

\title{Interactive Elicitation of Knowledge on Feature Relevance Improves Predictions in Small Data Sets}


\numberofauthors{1}
\author{
\alignauthor{Luana Micallef*$^{,1}$, Iiris Sundin*$^{,1}$, Pekka Marttinen*$^{,1}$,
Muhammad Ammad-ud-din$^{1}$, \\ Tomi Peltola$^{1}$, Marta Soare$^{1}$, Giulio Jacucci$^{2}$, and Samuel Kaski$^{1}$}\\
\affaddr{
\normalsize$^{1}$Helsinki Institute for Information Technology HIIT, Department of Computer Science, Aalto University, Finland
\\
$^{2}$Helsinki Institute for Information Technology HIIT, Department of Computer Science, University of Helsinki, Finland
\\
 $^{*}$Authors contributed equally \qquad firstname.lastname@hiit.fi}
}


\maketitle

\begin{abstract}
Providing accurate predictions is challenging for machine learning algorithms when the number of features is larger than the number of samples in the data. Prior knowledge can improve machine learning models by indicating relevant variables and parameter values. Yet, this prior knowledge is often tacit and only available from domain experts. We present a novel approach that uses interactive visualization to elicit the tacit prior knowledge and uses it to improve the accuracy of prediction models. The main component of our approach is a user model that models the domain expert's knowledge of the relevance of different features for a prediction task. In particular, based on the expert's earlier input, the user model guides the selection of the features on which to elicit user's knowledge next. The results of a controlled user study show that the user model significantly improves prior knowledge elicitation and prediction accuracy, when predicting the relative citation counts of scientific documents in a specific domain.

\end{abstract}

\vspace{-.6em}


\vspace{-.7em}




\graphicspath{{figures/}}



\newcommand{\RRR}{\mathbb{R}}
\newcommand{\EEE}{\mathbb{E}}
\newcommand{\NNN}{\mathbb{N}}
\newcommand{\xx}{{\mathbf x}}
\newcommand{\yy}{{\mathbf y}}
\newcommand{\zz}{{\mathbf z}}
\newcommand{\dd}{{\mathbf d}}
\newcommand{\hh}{{\mathbf h}}
\newcommand{\ttt}{{\mathbf t}}
\newcommand{\jj}{{\mathbf j}}
\newcommand{\pp}{{\mathbf p}}
\newcommand{\qq}{{\mathbf q}}
\newcommand{\aaa}{{\mathbf a}}
\renewcommand{\ggg}{{\mathbf g}}
\newcommand{\ssss}{{\mathbf s}}
\newcommand{\bb}{{\mathbf b}}
\newcommand{\ee}{{\mathbf e}}
\newcommand{\cc}{{\mathbf c}}
\newcommand{\nn}{{\mathbf n}}
\newcommand{\mm}{{\mathbf m}}
\newcommand{\kk}{{\mathbf k}}
\renewcommand{\lll}{{\mathbf \ell}}
\newcommand{\rr}{{\mathbf r}}
\newcommand{\uu}{{\mathbf u}}
\newcommand{\vv}{{\mathbf v}}
\newcommand{\ff}{{\mathbf f}}
\newcommand{\fff}{{\mathbf f}}
\newcommand{\JJ}{{\mathbf J}}
\newcommand{\DD}{{\mathbf D}}
\newcommand{\BB}{{\mathbf B}}
\newcommand{\CC}{{\mathbf C}}
\newcommand{\PP}{{\mathbf P}}
\newcommand{\MM}{{\mathbf M}}
\newcommand{\QQ}{{\mathbf Q}}
\newcommand{\FF}{{\mathbf F}}
\newcommand{\TT}{{\mathbf T}}
\newcommand{\RR}{{\mathbf R}}
\newcommand{\SSS}{{\mathbf S}}
\newcommand{\Sa}{{\mathbf {Sa}}}
\newcommand{\ZZ}{Z}
\newcommand{\cs}{{\dot c}}
\newcommand{\sss}{{\dot s}}
\newcommand{\ps}{{\dot p}}
\newcommand{\pg}{{\dot g}}
\newcommand{\ww}{{\mathbf w}}
\newcommand{\xxs}{\dot{\mathbf x}}
\newcommand{\vvs}{\dot{\mathbf v}}
\newcommand{\us}{{\dot u}}
\newcommand{\vs}{{\dot v}}
\newcommand{\ws}{{\dot w}}
\newcommand{\fn}{{\mathbf 0}}

\newcommand{\ii}{{\mathbf i}}
\newcommand{\Dt}{{\widetilde D}}
\newcommand{\fft}{{\widetilde \ff}}
\newcommand{\kkt}{{\widetilde \kk}}
\newcommand{\ggt}{{\widetilde \ggg}}
\newcommand{\hht}{{\widetilde \hh}}
\newcommand{\vvt}{{\widetilde \vv}}
\newcommand{\wwt}{{\widetilde \ww}}

\newcommand{\dbar}[1]{\bar{\bar{#1}}}
\newcommand{\qqqq}{\dbar{\qq}}

\newcommand{\eet}{\dbar{\ee}_{t}}

\newcommand{\eetau}{\dbar{\ee}_{\tau}}
%

\newcommand{\mNN}{\mathcal N}
\newcommand{\mEE}{\mathcal E}
\newcommand{\mSS}{\mathcal S}
\newcommand{\mRR}{\mathcal R}
\newcommand{\mPP}{\mathcal P}
\newcommand{\mPs}{\mathcal P_{S}}
\newcommand{\mBB}{\mathcal B}
\newcommand{\mDD}{\mathcal D}
\newcommand{\mOO}{\mathcal O}
\newcommand{\mAA}{\mathcal A}
\newcommand{\mWW}{\mathcal W}
\newcommand{\mMM}{\mathcal M}
\newcommand{\mXX}{\mathcal X}

\newcommand{\bmu}{{\bar{\mu}}}
\newcommand{\bsigma}{{\bar{\sigma}}}

\newcommand{\smallfrac}[2]{\frac{#1}{#2}}

\newcommand{\magn}[1]{|#1|}
\newcommand{\norm}[1]{\|#1\|}

\newcommand{\argmin}{\mathop{\textrm{arg\,min}}}

%
%


%
%



\newcommand{\todo}[1]{\warn\textbf{\small\textcolor{red}{TODO: #1}}}
\newcommand{\alternative}[2]{\warn{\footnotesize XXX \textbf{Alternative 1:} \emph{#1} \textbf{Alternative 2:} \emph{#2} XXX}}

\newcommand{\annoauthor}[3]{\warn\textbf{\small\textcolor[rgb]{#1}{#2: \emph{#3}}}}

\newcommand{\luana}[1] {\annoauthor{0.2,0.8,0.2}{Luana}{#1}}
\newcommand{\gregorio}[1]{\annoauthor{0.8,0.2,0.2}{Gregorio}{#1}}
\newcommand{\antti}[1] {\annoauthor{0.2,0.5,0.2}{Antti}{#1}}
\newcommand{\tino}[1] {\annoauthor{0.5,0.2,0.2}{Tino}{#1}}

\newcommand{\changed}[1]{#1}
\newcommand{\allchanged}{}
\newcommand{\allsame}{}

%
\newcommand{\compresslist}{
  \setlength{\itemsep}{2pt}
  \setlength{\parskip}{0pt}
  \setlength{\parsep}{0pt}
}
%

\makeatletter
    \renewcommand{\thesubfigure}{\alph{subfigure}}
    \renewcommand{\@thesubfigure}{\subcaplabelfont (\thesubfigure)\space}
    \renewcommand{\p@subfigure}{\thefigure}
\makeatother

\newlength{\lengthgoodgap}
\addtolength{\lengthgoodgap}{\subfigtopskip}
\addtolength{\lengthgoodgap}{\subfigbottomskip}
\newcommand{\goodgap}{\hspace{\lengthgoodgap}}
\newlength{\lengthlittlegap}
\addtolength{\lengthlittlegap}{\subfigtopskip}
\newcommand{\littlegap}{\hspace{\lengthlittlegap}}

\newlength{\twopicwidth}
\addtolength{\twopicwidth}{0.5\textwidth}
\addtolength{\twopicwidth}{-0.5\lengthgoodgap}

\newlength{\threepicwidth}
\addtolength{\threepicwidth}{0.333333\textwidth}
\addtolength{\threepicwidth}{-0.666666\lengthlittlegap}

\newlength{\fourpicwidth}
\addtolength{\fourpicwidth}{0.25\textwidth}
\addtolength{\fourpicwidth}{-0.75\lengthgoodgap}

\newlength{\fivepicwidth}
\addtolength{\fivepicwidth}{0.20\textwidth}
\addtolength{\fivepicwidth}{-0.80\lengthlittlegap}

\newlength{\sixpicwidth}
\addtolength{\sixpicwidth}{0.166666666666666667\textwidth}
\addtolength{\sixpicwidth}{-0.833333333333333333\lengthlittlegap}

\newlength{\ltwopicwidth}
\addtolength{\ltwopicwidth}{0.5\linewidth}
\addtolength{\ltwopicwidth}{-0.5\lengthgoodgap}

\newlength{\lthreepicwidth}
\addtolength{\lthreepicwidth}{0.333333\linewidth}
\addtolength{\lthreepicwidth}{-0.666666\lengthlittlegap}

\newlength{\lfourpicwidth}
\addtolength{\lfourpicwidth}{0.25\linewidth}
\addtolength{\lfourpicwidth}{-0.75\lengthlittlegap}

\ifx\hypersetup\undefined
	\newlength{\manualsubfigtopskip}
	\addtolength{\manualsubfigtopskip}{0cm}
\else
	\newlength{\manualsubfigtopskip}
	\addtolength{\manualsubfigtopskip}{\subfigtopskip}
	\setlength{\subfigtopskip}{0cm}
\fi

\hfuzz=3pt

\newcommand{\resetlength}[1]{\ifx#1\undefined \newlength{#1} \else \setlength{#1}{0pt} \fi}

%
%





\newcommand{\KayGrid}{
	\draw[style=help lines, color=gray, xstep=0.05, ystep=0.05] (0,0) grid (1, 1);%
	\draw[style=help lines, color=darkgray, xstep=0.1, ystep=0.1, thick] (0,0) grid (1, 1);%
	\draw[style=help lines, color=black, xstep=0.5, ystep=0.5, very thick] (0,0) grid (1, 1);%
}

\newcommand{\KayGridHalf}{
	\draw[style=help lines, color=gray, xstep=0.05, ystep=0.05] (0,0) grid (1, 0.5);%
	\draw[style=help lines, color=darkgray, xstep=0.1, ystep=0.1, thick] (0,0) grid (1, 0.5);%
	\draw[style=help lines, color=black, xstep=0.5, ystep=0.5, very thick] (0,0) grid (1, 0.5);%
}


\setlength{\abovedisplayskip}{4pt}
\setlength{\belowdisplayskip}{4pt}

\begin{figure*}[tb]
    \centering
    \subfigure[]{\label{fig:overview_ui} \includegraphics[width=0.33\textwidth]{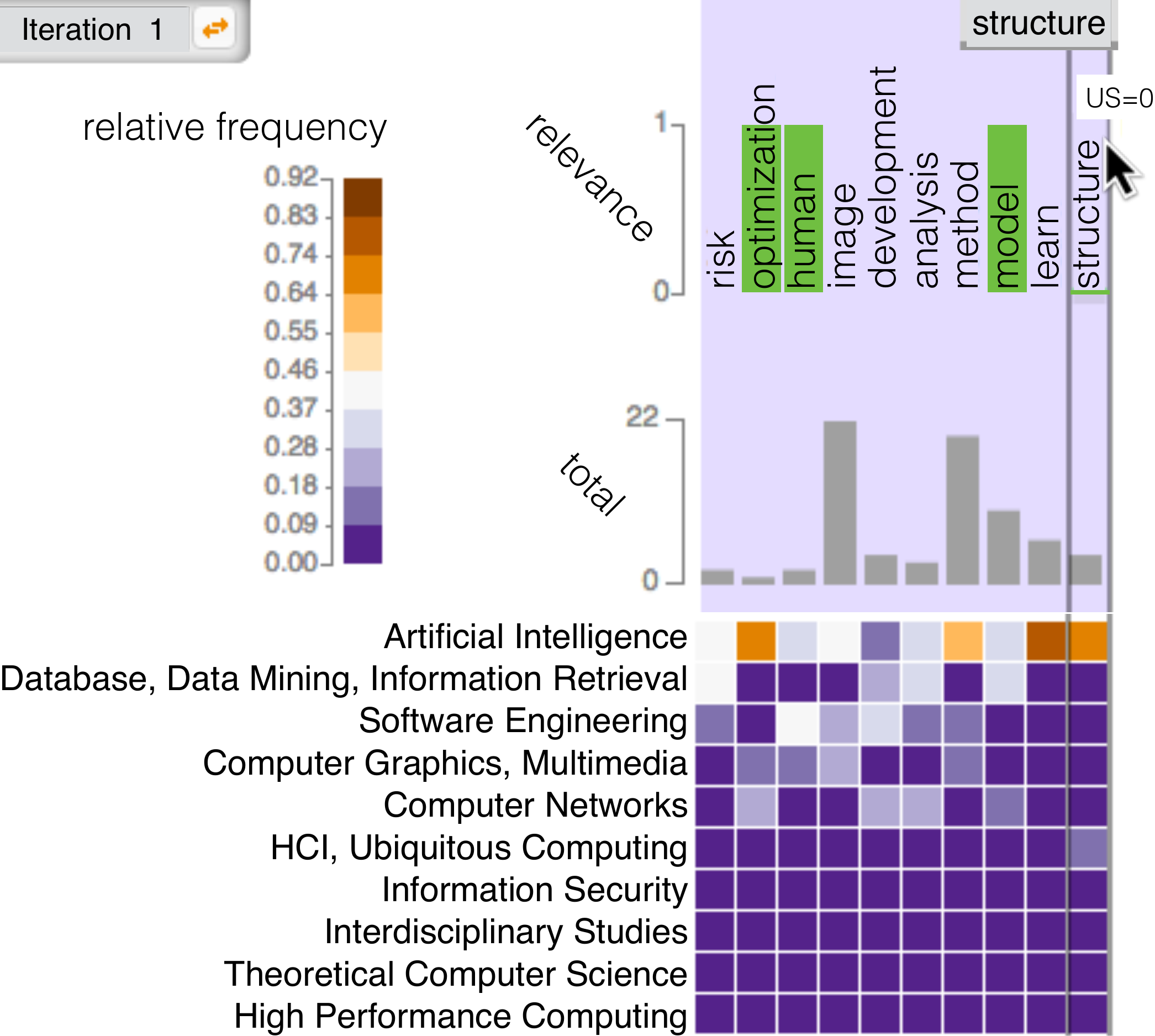}}
    \goodgap
    \subfigure[]{\label{fig:overview_flow} \includegraphics[width=0.63\textwidth]{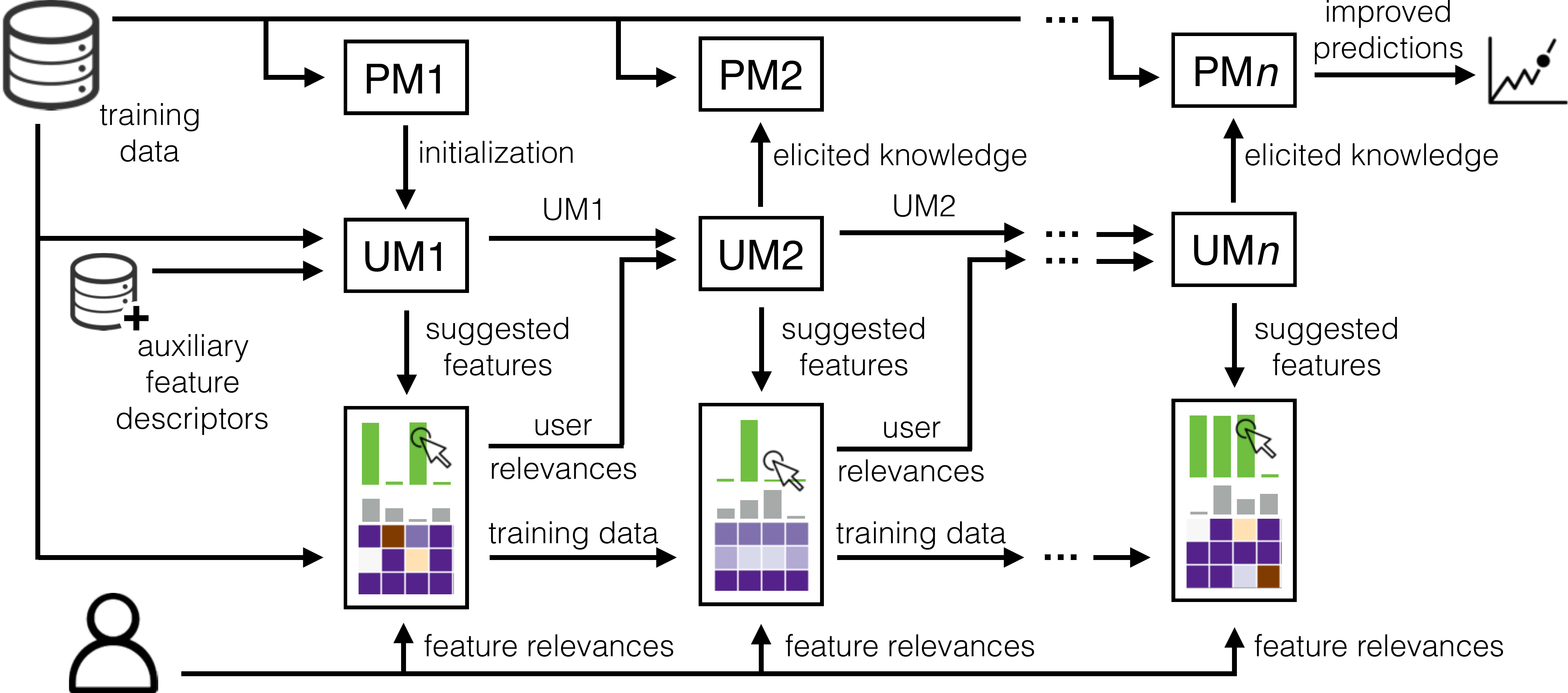}}
    \caption{Overview of our approach. (a) The user interface of an implementation of our approach. (b) Flow of data between the prediction model (PM), user model (UM) and interactive visualization at various iterations of the prior knowledge elicitation and prediction improvement process. See the `Method' section for details.}
    \label{fig:overview}
\end{figure*}

\vspace{-0.1em}
\section{Introduction}
We address the machine learning problem of predicting values of a target variable given a training data set in which the target variable values are known. The training data set needs to be representative of the underlying population, and its size must be large enough for the machine learning model to accurately learn to predict the \emph{target} variable. Yet, in applications like personalized medicine \cite{costello2014community,papillon2013comparison,tian2014simple}, brain imaging \cite{wang2011sparse,zhou2013tensor} and textual document categorization \cite{genkin2007large,kogan2009predicting,lewis1992feature,qu2010bag,zhang2006linear}, the number of features by far exceeds the number of samples, leading to the ``small $n$ large $p$'' problem \cite{friedman2001elements} where classical models inaccurately predict the target.
Fitting regression models for this problem requires regularizing the model's regression coefficients \cite{hoerl1970ridge,tibshirani1996regression,zou2005regularization}. Typically, the level of regularization is tuned by estimating a hyperparameter from the data, but this neglects prior information that could be available on the problem, the prior information referring to any knowledge of the problem the user may have before inspecting the data.
Yet, knowledge of the features' effects on the target could significantly improve predictions \cite{sinha2008incorporating}.

The use of prior knowledge in prediction is often not straightforward. For example, the prior information may not be available in any format that can easily be plugged into the prediction model. Nevertheless, a domain expert may possess tacit knowledge, not written down anywhere, of the relationships between the features and the target variable. Take, for example, the task of predicting the number of citations a scientific document receives in a certain domain. An expert can easily indicate that the presence of a term `neural' in the document implies a higher relative citation count in the machine learning domain. However, eliciting such tacit knowledge is difficult when the number of putative features is large, and checking each individual feature is excessively laborious.

We present a novel approach that extracts the tacit knowledge from the domain expert and uses this knowledge as prior information for improved predictions. 
A prediction model is still responsible for generating the predictions for the target variable. However, a user model selects features whose \emph{relevance} is indicated by the user, a domain expert, using an interactive visualization. Here, a \emph{relevant feature} is a feature that is positively correlated\footnote{correlated in general, even if not necessarily in the training data} with the target value. The user model iteratively elicits this information, to build a model of the user's tacit knowledge and select other features that would benefit from the user's input. The user input is then encoded into prior knowledge for the prediction model to improve its accuracy. 
Our contributions are:
\begin{itemize} \compresslist
\item We present a novel method that interactively models the user's tacit knowledge of the relevance of features to the predicted target, and uses this elicited information as prior knowledge for a more accurate prediction model.
\item Through a user study, we demonstrate that using a user model to select the features that require input from the domain expert significantly improves prior knowledge elicitation when compared to randomly selected features.
\end{itemize}

\section{Related Work}

Expert knowledge can be integrated into prediction models by defining prior distributions for model parameters. Typically, in prior elicitation full prior distributions have to be defined by experts \cite{garthwaite13prior,jones14prior,kadane80interactive}. This is time consuming and infeasible for high-dimensional problems, even with interactive tools. A simpler method for Bayesian Networks required experts to only indicate the presence or absence of the most uncertain causal relationships \cite{Cano11a}. In information retrieval, interactive intent modeling finds relevant resources based on user's previous input \cite{ruotsalo15interactive}. Deciding which features to ask user input on is done iteratively, by balancing the \textit{exploitation} of the currently most promising features and the \textit{exploration} of uncertain, possibly interesting ones. The balancing is done with linear bandit algorithms~\cite{auer2003using}.

Previously, interactive visualization has been used in classification tasks \cite{alsallakh2014visual,kapoor2010interactive,may2011guiding}. However, the underlying classification model itself is not directly modified, or the approaches are limited to cases with more samples than features. In \cite{brooks2015featureinsight}, possibly important features were visualized to the user and included interactively to a classifier, and in \cite{ribeiro2016why} the user was shown features that best explained predictions of a classifier, allowing her to reject irrelevant features. Semi-supervised clustering was considered in \cite{must-link-cannot-link}, where users indicated which pairs of items should belong to the same cluster.
However, simply including or excluding a feature is sensitive to errors and not sufficient in ``small $n$ large $p$'' problems. The method in \cite{soare16regression} tackles this problem with the simplifying assumption that the expert may give noisy input directly on the regression coefficients, and ~\cite{OLeary09} performed non-interactively a direct elicitation of logistic regression coefficients. In recent works considering a similar problem, a user specified the similarity of features as input \cite{afrabandpey2016interactive}, or features were chosen based on information gain \cite{daee2016knowledge}.

Our new approach for interactive visualization has the purpose of knowledge elicitation to improve the accuracy of a prediction model. Out approach differs from the methods above by using a 'user model' which adaptively learns the domain user's expert knowledge. It automatically guides the interaction towards features that would likely benefit from the user's input, based on the current representation of the expert's knowledge. Furthermore, the user model can exploit not only the training data, but also any additional auxiliary data about the features, important in scaling the method to small data sets.

\vspace{-0.4em}
\section{Method}
Fig.~\ref{fig:overview_flow} shows the main components of our approach, namely: the \emph{prediction model} (PM), the \emph{user model} (UM), and the \emph{interactive visualization} (IVis). An implementation of our approach is shown in Fig.~\ref{fig:overview_ui}.
IVis displays the training data and some features for which the user (a domain expert) has to indicate their relevance for a particular prediction task. UM then models the user's knowledge of feature relevances, and PM uses the user input with the training data to improve the predictions. The \emph{training data} (TD) is a small set of samples with a large number of features and the target. Additional data, referred to as \emph{auxiliary feature descriptors} (AFD), are required to provide information about the features that is not available in the training data. The flow of events in our approach is as follows:

\begin{enumerate}\compresslist
\item \textit{Initialize}. PM is initialized by TD. UM is initialized by TD, AFD, and information from the learned PM.
\end{enumerate}

Repeat 

\begin{enumerate}[resume]\compresslist
\item \textit{Select features to show}. UM is used to select a set of features to show next to the user.

\item \textit{Get user's input}. The user indicates the relevances of the shown features for the given prediction task, based on her prior tacit knowledge.

\item \textit{Update models}. UM and PM are updated using the relevances of the features provided by the user.
\end{enumerate}

Until ready

\begin{enumerate}[resume]\compresslist
\item \textit{Return predictions}. PM returns improved predictions.
\end{enumerate}
We briefly discuss each component below; details are provided in the Supplementary Material.

\subsection{Prediction Model}
We introduce the idea on a scalar-valued prediction problem with linear models, but the approach can be generalized.

\begin{figure}[hb]
    \centering
    \includegraphics[width=0.49\textwidth]{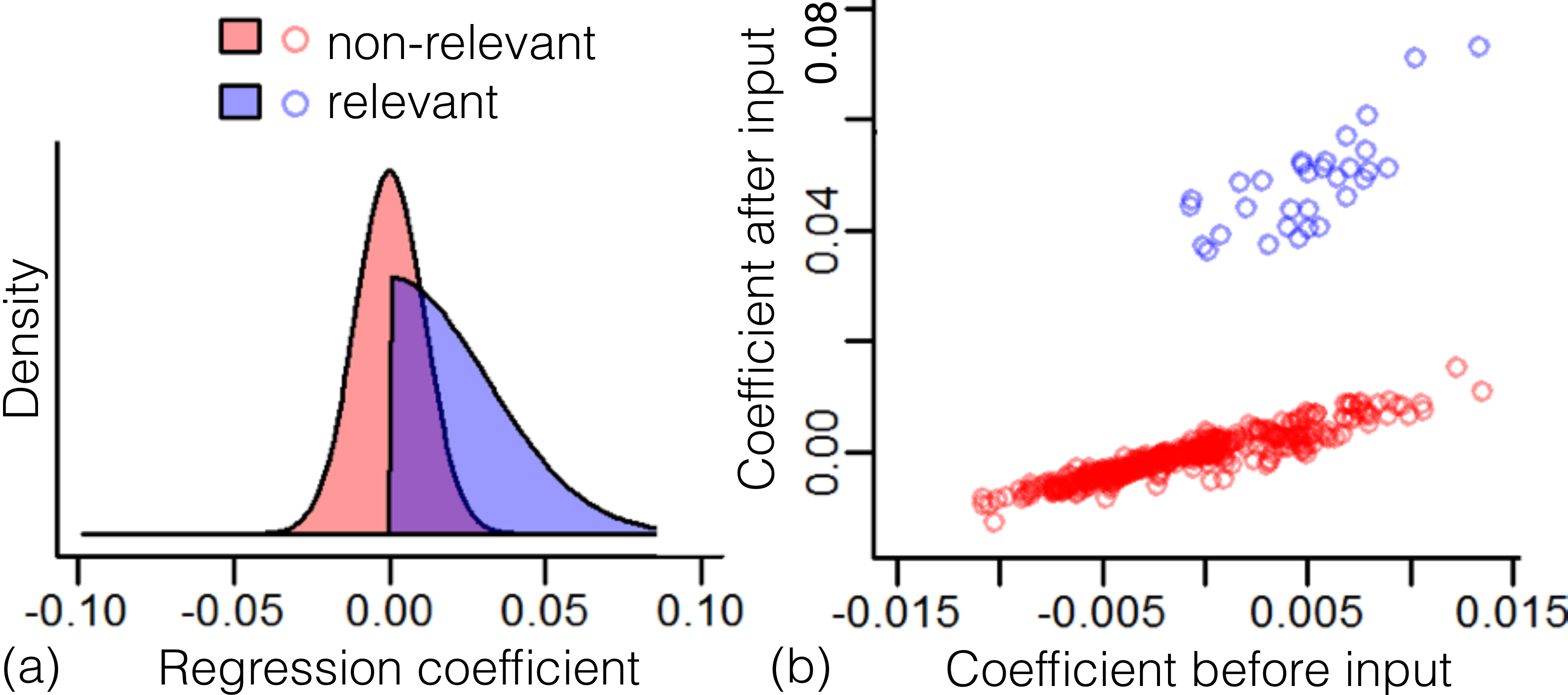}
    \caption{Regression coefficients of relevant and non-relevant features: (a) Prior distributions;  (b) Impact of user input.}
    \label{fig:predmodel}
\end{figure}

As input, the prediction model takes the training data points $(\mathbf{x}_{i},y_{i}),~i=1,\ldots,N$, where $\mathbf{x}_{i}\in\mathbb{R}^{K}$ are the features and $y_{i}\in\mathbb{R}$ the value of the target variable for sample $i$. In addition, a vector of relevances $\mathbf{r}\in\{0,1\}^{K}$ is provided, where $r_{j}=1$ if the feature is relevant, i.e., has received positive user input, and $r_{j}=0$ otherwise. We assume a linear prediction model
\begin{equation*}
y_{i}\sim N(\mathbf{x}_{i}^{T}\mathbf{w},\sigma^{2}),\text{ }i=1,\ldots,N,
\end{equation*}
where $\mathbf{w\in}\mathbb{R}^{K}$ is a vector of regression coefficients and $\sigma^{2}$ the variance of the Gaussian noise. The relevances of the features $\mathbf{r}$ enter the prediction model through modifying the prior distribution of the elements of $\mathbf{w}$ as follows:
\begin{align*}
w_{j}  & \sim N(0,\sigma_{0}^{2}),\text{ if }r_{j}=0,\\
w_{j}  & \sim\text{half-}N(0,a\sigma_{0}^{2}),\text{ if }r_{j}=1.
\end{align*}
Half-$N$ denotes the half-normal distribution. The intuition is that if a feature is deemed relevant, its presence is assumed to increase the value of the output variable (Fig.~\ref{fig:predmodel}a). The multiplier $a$ determines the overall ratio of the effect sizes between relevant and non-relevant features. Fig.~\ref{fig:predmodel}b shows the impact of this formulation on the estimated regression coefficients. 

\subsection{User Model}
Efficient interaction balances between querying additional input on either the most promising relevant features (\textit{exploitation}), or on the most uncertain ones (\textit{exploration}). This is achieved by using the upper confidence bound criterion (UCB) to select features to show to the user, as in the algorithm LINREL \cite{auer2003using}. At each iteration $t$, a user is shown $n_t$ features with highest UCBs from the previous iteration. The user then specifies a binary relevance $r_j \in \{0,1\}$ value to each feature $j$. At each iteration, the user model updates the estimated feature relevances $\hat{r}_{j,t}$ using a linear model:
\begin{equation*}
    \hat{r}_{j,t} = Z_j\hat{\mathbf{v}}_t+b \quad \forall \quad j \in 1,\ldots,K
\end{equation*}
where $Z_j \in \mathbb{R}^{N_Z}$ is a feature descriptor of the $j$th feature and $b$ determines the default relevance. The $\hat{\mathbf{v}}_t$ is a vector of regression coefficients, and it is estimated from inputs given so far, using the standard regularized least squares solution. The relevances are converted to interval $(0,1)$ using the logistic transformation.

Feature descriptors $Z_j$ are chosen depending on the problem domain, and they can be constructed from the training data and/or any auxiliary data in which the features, but not necessarily the target variable, are available. For example, in the evaluation study, we use the \textit{tf-idf} \cite{jones72astatistical} of keywords in clusters of scientific documents. The intuition is that keywords that appear in similar documents have similar effect in the prediction task, and should thus have correlated feature descriptors.
Finally, the UCBs are defined as $r_{j,t}^{UCB} =\hat{r}_{j,t} + c_{j,t}$, where $c_{j,t} $ is a high probability bound for relevance uncertainty, computed using SupLinUCB in \cite{chu2011contextual}.

\subsection{Interactive Visualization}
A \emph{heatmap} using a color-blind safe color scale\footnote{obtained from http://colorbrewer2.org} depicts the training data (Fig.~\ref{fig:overview_ui}). Rows indicate categories to which the samples are grouped (e.g., domains in which scientific documents were cited). Columns indicate features selected by the user model for which user input is required (e.g., words in a document). The cell color indicates how strongly, on average, the feature was associated to samples in that category (e.g., the average relative citation count in that domain for documents containing the word), with \emph{total bars} (in grey) above the heatmap showing the total number of samples on which this value was based, to get an idea of the reliability of the training data. By clicking on the feature labels, the user can set \emph{relevance bars} (in green) to either 1 or 0, indicating whether that feature is respectively relevant or not to the predicted target (e.g., being cited in the Artificial Intelligence domain). The relevance bars provide the domain expert the means to input her tacit knowledge. Even though the heatmap and the total bars showing the training data could help the domain expert decide the feature relevances, they are not essential for our approach. Nonetheless, we still evaluated their usefulness through a post-questionnaire in our user study (see the `Results and Discussion' section). 
\section{Evaluation}
We conducted computational and empirical experiments to evaluate our approach in a real-world scenario. 

The experiment conditions included: 
\begin{itemize}\compresslist
    \item \textbf{C1}: \emph{non-interactive} prediction model;
    \item \textbf{C2}: \emph{interactive} prediction model with features for user input suggested \emph{randomly};
    \item \textbf{C3}: \emph{interactive} prediction model with features for user input suggested by the \emph{user model}.
\end{itemize}
The task was to predict the relative citation count a scientific document will get in the domain of Artificial Intelligence (target variable) given that it has certain words (features) in the title, abstract or keywords. In C2 and C3, participants had to indicate whether each of the 10 suggested features were relevant or not to the target, for 20 iterations.

The data we used was a subset of Tang et al.'s citation data set \cite{tang2008arnetminer} containing 162 scientific documents, for which we: (i) manually retrieved the author provided keywords; (ii) automatically extracted additional keywords from the title and abstract of the documents using Python Rake \cite{rose2010automatic} and KP-Miner \cite{el2009kp}; (iii) lemmatized all the keywords obtained in \textit{i} and \textit{ii} using Python Natural Language Toolkit \cite{bird2006nltk}. This resulted in 457 unique keywords that were used as features. The data collection was evenly split into a training set and a test set. The training set was used to train the prediction model in C1-C3, while the test set was used to evaluate the accuracy of the predictions, using the Mean Squared Error (MSE).

Our hypotheses were: 
\begin{itemize}\compresslist
    \item \textbf{H1}: C2 and C3 provide more accurate predictions than C1;
    \item \textbf{H2}: C3 provides more accurate predictions than C2.
\end{itemize}

We adopted a between-participant design: 12 participants for C2 (8 males); 11 participants for C3 (9 males)\footnote{11 not 12 as the results of one participant were discarded as s/he provided incorrect input to the words learned in the training phase}. All participants: had at least 2 years research experience in machine learning; were undertaking a PhD or postdoc (1st or 2nd year PhD: 4 in C2, 3 in C3); were at least somewhat familiar with heatmaps and bar charts; were aged 20-40. 
Each participant was trained to use the system (Fig.~\ref{fig:overview_ui}), introduced to the prediction task, and asked to complete the task for one iteration. The answers were then discussed with the experimenter and the participant was given 10 more min to explore the system before the actual experiment. 
At the end, participants filled in a questionnaire. The experiment took $\approx$30mins and a movie ticket was awarded. For details, see Supplementary Material.
\section{Results and Discussion}

\begin{figure}[ht]
    \centering
    \hspace*{-.5cm}    
    \includegraphics[width=0.49\textwidth]{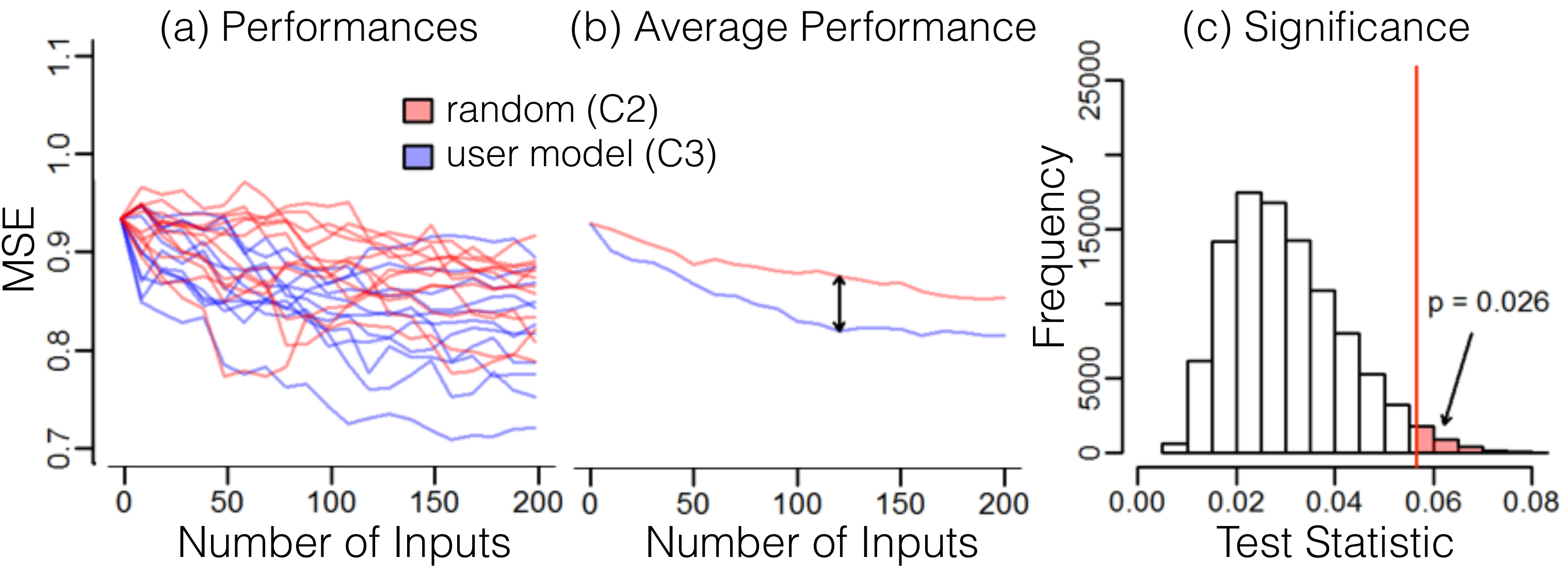}
    \caption{a) Mean squared error w.r.t. the number of inputs provided by our study participants. b) Test statistic is the maximum distance between average performance curves. c) Distribution of the test statistic in $10^6$ permutations.}
    \label{fig:results}
    \vspace{-0.4em}
\end{figure}

The final predictions of C2 and C3 were more accurate than those of C1 for all 23 participants, i.e., user input always increased prediction accuracy, and the Mean Squared Error (MSE) decreased as the participants provided more input (Fig.~\ref{fig:results}a). MSE without user input (C1) was 0.93, and with user input (C2 and C3) after the interaction 0.84 (mean) $\pm$0.05 (sd). Average performance at the end is significantly different from performance without user input (\emph{p}=2.3e-7, Wilcoxon signed-rank test), confirming H1. Thus, without user input the prediction model explained about 7\% of the variance in the target variable, and with user input 16\%.

To evaluate the difference between giving user input in a random order (C2) vs. with the user model (C3), we computed the average MSE curves in the two groups (Fig.~\ref{fig:results}b). The random order improves predictions approximately linearly w.r.t. the number of user input, whereas with the user model the predictions improve more rapidly at early stages of interaction, as expected. We used the maximum distance between the average curves as the test statistic to characterize this difference (Fig. \ref{fig:results}b). We computed the distribution of the test statistic, assuming no difference between groups, using $10^6$ permutations of the group labels (Fig. \ref{fig:results}c), which shows that the difference is significant (\emph{p}=0.026), thus confirming H2.

The results of the post-questionnaire (Table \ref{tab:table1}) indicate that: (i) the visualization of the training data (heatmap+total bars) is used more when the user is uncertain about the feature relevance (as in C2); (ii) when the heatmap is referred to, the total bars are more carefully analysed to verify the reliability of the displayed data (as in C2); and (iii) the visualization is familiar and simple enough for a domain expert to understand and use. In summary, these findings suggest that visualizing the data is useful when eliciting expert feedback, inspiring us to develop the visualization further in the future.


In our approach we query the user about whether a feature is relevant, i.e., is positively correlated with the target variable. This is a compromise between detailed input about regression coefficients (exact value \cite{soare16regression} or full prior \cite{garthwaite13prior,jones14prior} and simple input discarding a subset of features \cite{Cano11a,ribeiro2016why}). This kind of user input is easy to give 
(difficulty C2 and C3, self-reported in post-study survey: 50\% easy, 29\% neutral), but powerful in improving the predictive performance. However, the model is potentially sensitive to errors in user input. Also, although providing user input on positive effects was natural for the prediction task considered here, in other cases negative user input may be useful. We will consider these issues further in future work.
Our user model formulation has the additional benefit of allowing integration of auxiliary data when defining feature descriptors. This is particularly important when the sample size decreases, and training data alone would not provide enough information to guide user interaction.

\begin{table}
  \centering
  \begin{tabular}{l r r}
    & \small{\textbf{Random (C2)}} & \small{\textbf{User model (C3)}} \\
    \midrule
    Referred to the heatmap & 91.7\% & 75\% \\
    Found the heatmap helpful & 58.4\% & 50\% \\
    \midrule
    Referred to the total bars & 58.3\% & 25\% \\
    Found the total bars helpful & 41.7\% & 16.6\% \\
    \midrule
    Found the words relevant & 66.7\% & 75\% \\
    \midrule
    Confident with their answers & 75\% & 83.3\% \\
    Were not at all confident & 8.3\% & 0\% \\
  \end{tabular}
  \caption{Results of the post-questionnaire (N=23)
  }~\label{tab:table1}
  \vspace{-0.7cm}
\end{table}
 \section{Conclusion}
We have presented a novel approach for eliciting tacit knowledge from domain experts and using it as prior knowledge to improve the accuracy of prediction models for ``small $n$ large $p$'' problems. A user study indicates the effectiveness of our approach in contrast to a non-interactive prediction model, and one that is interactive but suggests features for user input at random. In the future, we will: evaluate this approach on other real-word data; explore how visualizations can facilitate knowledge elicitation; and investigate ways how  to extend the prediction model to multiple output learning.


\section{Acknowledgments}
We thank the participants of our user study. We acknowledge the computational resources provided by the Aalto Science-IT project. This work was funded by the Academy of Finland (Centre of Excellence in Computational Inference Research COIN, and grants no. 295503, 294238, 286607 and 294015), and by Jenny and Antti Wihuri Foundation.

%
%
%
%
%
\balance

\bibliographystyle{acm-sigchi}
\bibliography{references}

\pagebreak[1]

\onecolumn

\graphicspath{{figuressupp/}}

\begin{center}
Supplementary Material
\end{center}

\section*{\centering \large{\textbf{Interactive Elicitation of Knowledge on Feature Relevance Improves Predictions in Small Data Sets}}}
\subsection*{
\centering{
\normalsize Luana Micallef*$^{,1}$, Iiris Sundin*$^{,1}$, Pekka Marttinen*$^{,1}$,
Muhammad Ammad-ud-din$^{1}$, Tomi Peltola$^{1}$, Marta Soare$^{1}$, Giulio Jacucci$^{2}$, and Samuel Kaski$^{1}$}}
\vspace{4mm}
\begin{center}
{\small{
$^{1}$Helsinki Institute for Information Technology HIIT,\\ Department of Computer Science, Aalto University, Finland
\\
$^{2}$Helsinki Institute for Information Technology HIIT,\\ Department of Computer Science, University of Helsinki, Finland
\\
 $^{*}$Authors contributed equally \\ firstname.lastname@hiit.fi\\
 }}
\end{center}


\vspace{.3in}

\noindent
This supplementary provides addition information about the prediction model (Section~\ref{sec:supp_predictionmodel}), the user model (Section~\ref{sec:supp_usermodel}), and our evaluation (Section~\ref{sec:supp_evaluation}).

\section{Prediction Model}
\label{sec:supp_predictionmodel}
As input, the prediction model takes training data points $(\mathbf{x}_{i},y_{i}),~i=1,\ldots,N,$ where $\mathbf{x}_{i}\in\mathbb{R}^{K}$ and $y_{i}\in\mathbb{R}$, and a vector of relevances $\mathbf{r}\in\{0,1\}^{K},$where $r_{j}=1,$ if the feature is relevant, i.e., has received positive feedback. Otherwise, $r_{j}=0$.

\begin{figure}[h]
    \centering
    \includegraphics[width=\textwidth]{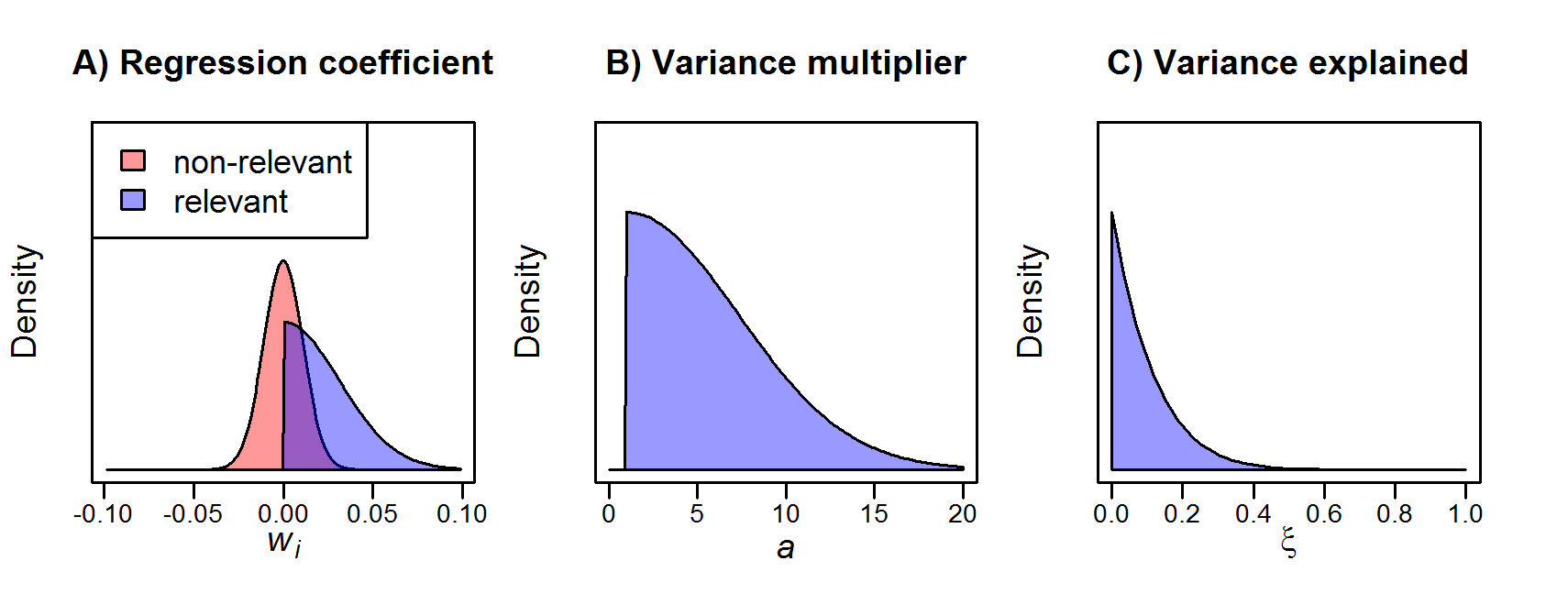}
    \caption{Prior distributions of the parameters of the prediction model}
    \label{fig:priors}
\end{figure}

We assume the target $y_{i}$ depends linearly on the predictor $\mathbf{x}_{i}$%
\[
y_{i}\sim N(\mathbf{x}_{i}^{T}\mathbf{w},\sigma^{2}),\text{ }i=1\ldots,N,
\]
where $\mathbf{w\in}\mathbb{R}^{K}$ is a vector of regression coefficients and $\sigma^{2}$ is the variance of the Gaussian noise. The relevances of the predictors $\mathbf{r}$ affect the prior distributions of the elements of $\mathbf{w}$ as follows%
\begin{align*}
w_{j}  & \sim N(0,\sigma_{0}^{2}),\text{ if }r_{j}=0,\\
w_{j}  & \sim\text{half-}N(0,a\sigma_{0}^{2}),\text{ if }r_{j}=1.
\end{align*}
Here, half-$N$ denotes the half-normal distribution. The intuition is that if a feature is relevant, the corresponding regression weight is assumed to have a prior distribution constrained to be positive (see, Fig. \ref{fig:priors}A). The multiplier $a$ determines the ratio of the variance parameters between relevant and non-relevant features, and is given a prior distribution
\[
a\sim1+\text{half-}N(0,12.5\pi).
\]
This constrains $a$ to be greater than $1$ and have mean 6, according to a weakly informative prior (see Fig. \ref{fig:priors}B). This corresponds to the expectation that regression coefficients of the relevant features are greater in magnitude than the coefficients of the non-relevant features.

The term $\sigma_{0}^{2}$ appearing in the prior variances of the regression
coefficients of both relevant and non-relevant features is specified by
investigating the variance of the linear predictions. A direct integration of
regression weights $\mathbf{w}$, conditional on parameters $a$ and $\sigma
_{0}^{2}$, gives%
\begin{align}
E_{\mathbf{w}}\left[  Var(\mathbf{x}^{T}\mathbf{w)}\right]   &  =\frac{1}{N}%
{\textstyle\sum\nolimits_{i}}
\mathbf{x}_{i}^{T}E_{\mathbf{w}}\left[  \mathbf{ww}^{T}\right]  \mathbf{x}%
_{i}\nonumber\\
&  =\sigma_{0}^{2}(n_{-}+an_{+})+\frac{2a\sigma_{0}^{2}}{\pi}\sum_{k\in R}%
\sum_{\substack{h\in R\\h\neq k}}\sigma_{kh}\label{two_terms}\\%
&  \approx\sigma_{0}^{2}(n_{-}+an_{+}),\label{approximation}%
\end{align}
where $n_{+}$ and $n_{-}$ are the numbers of relevant and non-relevant
features, $R$ is the set of all relevant features, and $\sigma_{kh}$ is the
covariance between the $k^{th}$ and $h^{th}$ features. In practice the second
term in Equation \ref{two_terms} is less than 25\% of the first term, and
therefore, we retain only the first term to keep the computations simple (this
is exact when the relevant features are uncorrelated). Let
$\xi$ denote the proportion of variance explained by the prediction model.
Assuming $y$ is normalized, the proportion of variance explained is given by Equation \ref{approximation}, and we can solve for $\sigma_{0}^{2}$ for any $\xi$ by
using:%
\[
\sigma_{0}^{2}(n_{-}+an_{+})=\xi,
\]
which yields%
\begin{equation}
\sigma_{0}^{2}=\frac{\xi}{n_{-}+an_{+}}.\label{sigma_0}%
\end{equation}
We define a prior for $\xi$ as
\[
\xi\sim Beta(1,9),
\]
shown in Fig. \ref{fig:priors}C, which corresponds to the expectation that
approximately 10\% of the variance of the target is explained by the
prediction model. This further imposes a prior on $\sigma_{0}^{2}$ through
Equation \ref{sigma_0}. Finally, we place the following prior on noise
variance%
\[
\sigma\sim\text{half-}N(0,1),
\]
which completes the definition of the prediction model. The model is implemented using the probabilistic programming language Stan \cite{carpenter16stan}.

\section{User Model}
\label{sec:supp_usermodel}
Efficient interaction balances between querying additional input on either the most promising relevant features (\textit{exploitation}), or on the most uncertain ones (\textit{exploration}). The upper confidence bound criterion (UCB) to select features to show to the user achieves this, as in the algorithm LINREL \cite{auer2003using}. At each iteration $t$, a user is shown $n_t$ features with the highest UCBs from the previous iteration. The user then specifies a binary relevance $r_j \in \{0,1\}$ value to each feature $j$. We denote the inputs collected from the user before or at iteration $t$ by $\mathbf{r}_t \in \mathbb{R}^{\sum_{i=1}^t n_i}$. At each iteration, the user model updates the estimated feature relevances $\hat{r}_{j,t}$ using a linear model:
\begin{equation*}
    \hat{r}_{j,t} = Z_j\hat{\mathbf{v}}_t+b \quad \forall \quad j \in 1,\ldots,K
\end{equation*}
where $Z_j \in \mathbb{R}^{N_Z}$ is a feature descriptor of the $j$th feature and $b$ determines the default relevance. $\hat{\mathbf{v}}_t$ is a vector of regression coefficients, and it is estimated from inputs given so far, using a standard formula for regularized regression:
\begin{equation*}
    \hat{\mathbf{v}}_t = (Z_t^\top Z_t + \lambda \mathbf{I})^{-1} Z_t^\top (\mathbf{r}_t-b),
\end{equation*}
where $\lambda$ is a regularizer, $Z \in \mathbb{R}^{K,N_Z}$ a feature descriptor matrix, and its sub-matrix $Z_t$ contains the descriptors corresponding to features that have received user input thus far. Furthermore, we convert the relevances to the interval $(0,1)$ using the logistic transformation.

A high probability bound, $c_{j,t} $, for the relevance uncertainty $P\left( |\tilde{r}_j - \hat{r}_{j,t}| \leqslant c_{j,t} \right) \leqslant 1-\delta$, where $\tilde{r}_j$ is the true relevance, can be derived using SupLinUCB \cite{chu2011contextual}: 
\begin{equation*}
    c_{j,t} =\rho_t \sqrt{Z_j^\top(Z_tZ_t^\top + \lambda \mathbf{I})^{-1}Z_j}, \hspace{0.4cm}
     \rho_t = \sqrt{\alpha\log \big(\frac{2tK}{\delta}\big)}. 
\end{equation*}
The UCBs are then defined as
\begin{align*}
     r_{j,t}^{UCB} =\hat{r}_{j,t} + c_{j,t}.
\end{align*}
The parameter $\alpha$ determines the exploration-exploitation trade-off. In the Evaluation, $n_t=10 \quad \forall \quad t \in 1,\ldots,20$, $b$=0.5, $\lambda$=1e-3, $\alpha$=0.5 and $\delta$=0.05. The user model selects the features with the largest UCBs that have not yet been selected, to avoid querying the same feature twice.


\subsubsection*{Initialization} 
We initialize the user model with pseudo-input in order to choose as relevant first 10 features as possible. We use the feature's regression coefficient $w_j$ from the non-interactive prediction model as pseudo-input, since the input to the features with the highest regression coefficients has the greatest potential in improving the predictions \cite{soare16regression}. The impact of pseudo-input is set to be weak, so that 10 pseudo-inputs correspond to one real user input. Therefore the impact of pseudo-feedback decreases as more user input is received.

Pseudo-input can be included in $\mathbf{r_t}$, or, if expressed explicitly as $\mathbf{r_0}$, the regression coefficients are
\begin{equation*}
    \hat{\mathbf{v}}_t = (Z_t^\top Z_t + \beta Z^\top Z + \lambda \mathbf{I})^{-1} (Z_t^\top (\mathbf{r}_t-b) + \beta Z^\top (\mathbf{r_0}-b)),
\end{equation*}
where $Z$ contains the feature descriptors of all features, and $\beta$=0.01 defines the strength of the pseudo-input.

\subsubsection*{Feature Descriptors in Evaluation}
For the evaluation study, we use \textit{tf-idf} \cite{jones72astatistical} of words in clusters of scientific documents as feature descriptors.  The intuition is that words that appear in similar documents have similar effect in the prediction task, and should thus have correlated feature descriptors. Furthermore, words that appear evenly in all clusters are likely not very useful for the prediction.

The feature descriptors $Z_j$ are constructed using auxiliary data on keywords from \cite{glowacka013directing}, in combination with our prediction data set. From the auxiliary data, only the documents that had at least one common keyword with the prediction data set were used. This results in 8554 unique documents with 26333 unique (lemmatized) keywords as features. We can use all data available on the features (in both training and test samples) because target variables are not used when constructing feature descriptors. In so doing we utilize maximal amount of information available without risking over-fitting the model.

The documents were clustered to 20 clusters by hierarchical clustering based on their cosine distance in the feature space. Randomly chosen 1000 documents were used to train the model, and the rest of the documents were assigned to clusters based on distance to cluster centers. This results in a feature descriptor matrix $Z \in \mathbb{R}^{K,20}$, where the element $z_{j,c}$ is the \textit{tf-idf} of a word $j$ in cluster $c$. The \textit{tf}-score of a word is computed cluster-wise, and \textit{idf} document-wise.

\section{Evaluation}
\label{sec:supp_evaluation}
The following are the documents provided to the participants during the controlled experiment for the training phase and the actual experiment task phase, and the questionnaire participants had to fill in at the end of the experiment together with its results. 



\section*{Supplementary material}

# Training

*Scientific document citations*

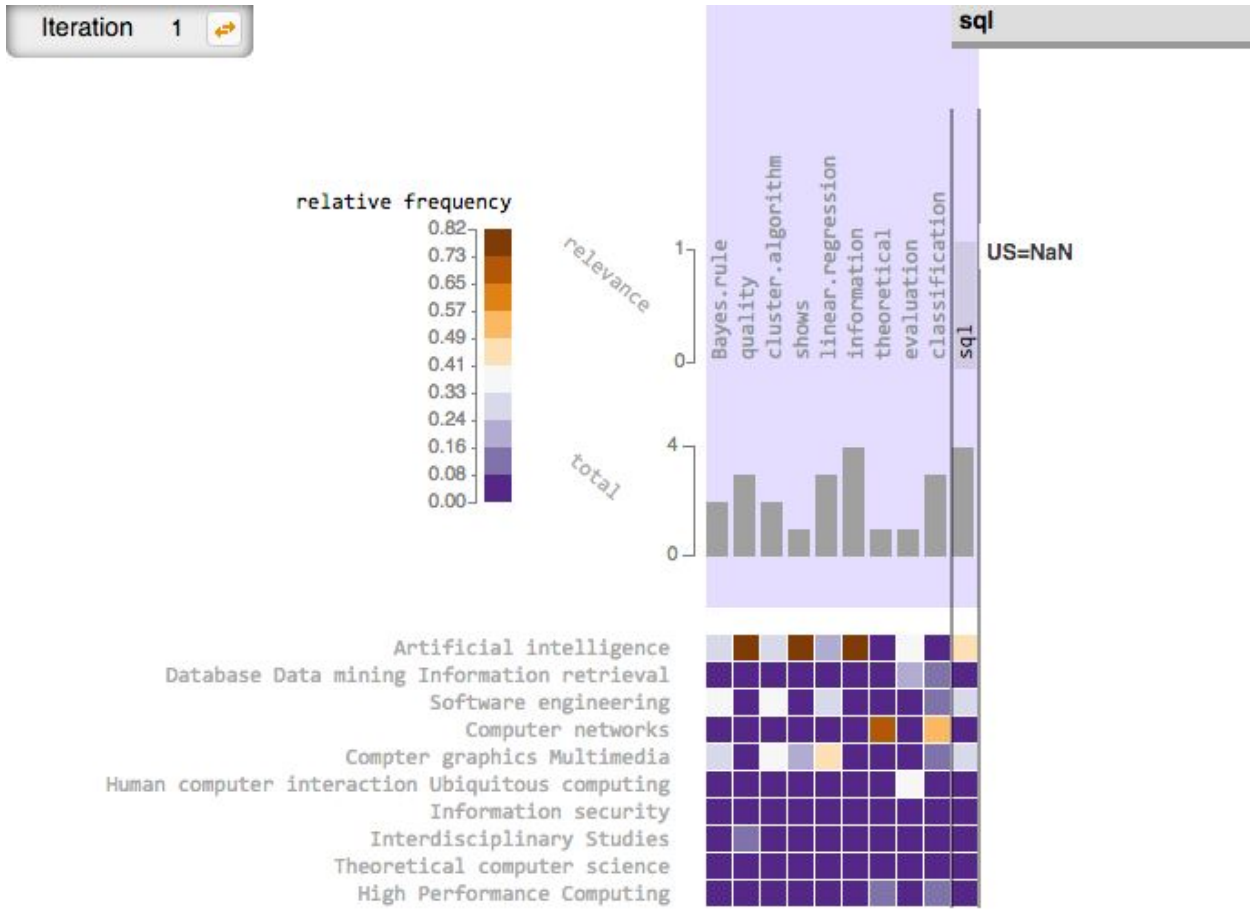

When the system loads, you will see something like the above.

The violet section shows a selection of **words obtained from the title, abstract and keywords of 81 scientific documents** from various journals. These scientific documents have been cited in one or more of the 10 domains listed above on the left of the heatmap matrix, such as *Artificial Intelligence*, *Software Engineering*, and *Computer networks*.

Looking at each of the 10 words in the row labelled *'relevance'*, your main task is to indicate **whether or not having that word** in a scientific document **increases the likelihood** that the document is **cited in the domain of Artificial Intelligence**.

In the image above, the user is currently with the cursor over the rightmost column titled '*sql*'.

- The tooltip '*US=NaN*' indicates that: the user has not given any input for the word '*sql*' as the user score (US) is still set to not a number (NaN).

- If the user thinks that '*sql*' **increases** the likelihood that a document is cited in the domain of Artificial Intelligence, as it is **representative** of this domain, then she should click on the **upper** half of the light greyish bar that appears **behind** '*sql*' to set **relevance to 1** (and US=1 in the tooltip).

  If the user thinks that '*sql*' **does NOT increase** the likelihood that a document is cited in the domain of Artificial Intelligence, as it is definitely **not representative** of this domain or it is **relevant to all other domains** not only Artificial Intelligence, then she should click on the **lower** half of the light greyish bar that appears **behind** '*sql*' to set **relevance to 0** (and US=0 in the tooltip).

  For instance,
    - words like '*quality*', '*information*' and *'sql'* **do not increase** the likelihood that a document is cited in the domain of Artificial Intelligence and the relevance of such words should be set to **0**, but
    - words like *'Bayes.rule'* and '*classification*' surely **increase** the likelihood that a document is cited in the domain of Artificial Intelligence and the relevance of such words should be set to **1**.

The *n*th **grey bar** in the row labelled '*total*' indicates

- the **number of scientific documents** (out of the 81 analysed documents) **containing the nth word** in the row labelled *'relevance'*

The **heatmap*** matrix below the violet section shows

- 10 different research **domains** as rows and 10 **words** as columns
  such that the cell (*r,w*) corresponding to the row for rating *r* and the column for word *w* indicates how much **on average** a scientific document containing the word *w* increases the likelihood that the document is cited in domain *d*

\* Data in the heatmap is based on ONLY a few examples of documents (81 documents in all), thus the depicted information might **NOT be a good representation** of the general case

You should now complete the following steps:

1. Look at the 10 words in the row labelled '*relevance*'. Start by focusing on the leftmost word in this row.

2. Move your cursor on the word you are currently focusing on in the row labelled '*relevance*', and click on the upper or lower half of the light greyish bar that appears behind that word to set its relevance score to 1 or 0, and thus respectively indicate whether that **word increases or does not increase the likelihood that a document is cited in the domain of Artificial Intelligence**. Use the heatmap matrix to help you make your decision.

3. Repeat step 2 for each of the 10 words in the row labelled '*relevance*'.

4. Click on the 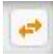 button.

5. Once a new set of 10 words appear in the row labelled '*relevance*', in the **actual experiment (not now)** you will have to repeat steps 1-4 other 19 more times (20 iterations in total). But now let the experimenter know you are done and ask any questions you have. Make sure you have understood both the task and the system before proceeding to the actual experiment.

# Experiment Task

*Scientific document citations*

Let the experimenter load the system and then complete the following steps as instructed in the training phase:

1. Look at the 10 words in the row labelled '*relevance*'. Start by focusing on the leftmost word in this row.

2. Move your cursor on the word you are currently focusing on in the row labelled '*relevance*', and click on the upper or lower half of the light greyish bar that appears behind that word to set its relevance score to 1 or 0, and thus respectively indicate whether that **word increases or does not increase the likelihood that a document is cited in the domain of Artificial Intelligence**. Use the heatmap matrix to help you make your decision.

3. Repeat step 2 for each of the 10 words in the row labelled '*relevance*'.

4. Click on the 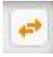 button.

5. Once a new set of 10 words appear in the row labelled '*relevance*', **repeat steps 1-4 another 19 more times** (20 iterations in total).

6. Let the experimenter know you are done, so she could download your results file (logs).

# User Study: Post Questionnaire

Fill out this questionnaire with respect to the **scientific document citations** experiment you have just completed.

* Required

1. **Experiment** *
   The experimenter (not participant) should complete this
   *Mark only one oval.*

   ◯ R
   ◯ U

2. **Participant's ID** *

3. **How FAMILIAR are you with heatmaps and bar charts, like the ones shown?** *
   *Mark only one oval.*

   ◯ not at all (never use them)
   ◯ slightly (almost never use them)
   ◯ somewhat (use them occasionally)
   ◯ moderately (use them moderately)
   ◯ extremely (use them frequently)

4. **Which aspects of the User Interface or Visualization (if any) did you find CONFUSING?** *

5. **Did you REFER TO the data shown in the HEATMAP?** *
   *Mark only one oval.*

   ◯ Never
   ◯ Almost never
   ◯ Occasionally/Sometimes
   ◯ Almost every time
   ◯ Frequently

6. **How HELPFUL did you find the data shown in the HEATMAP to complete the task?** *

   *Mark only one oval.*

   - ○ Not at all
   - ○ Slightly
   - ○ Somewhat
   - ○ Moderately
   - ○ Extremely

7. **How did you USE the data shown in HEATMAP to complete the task?** *

   _____
   _____
   _____
   _____
   _____

8. **Did you REFER TO the data shown in the TOTAL BARS?** *

   *Mark only one oval.*

   - ○ Never
   - ○ Almost never
   - ○ Occasionally/Sometimes
   - ○ Almost every time
   - ○ Frequently

9. **How HELPFUL did you find the TOTAL BARS?** *

   *Mark only one oval.*

   - ○ Not at all
   - ○ Slightly
   - ○ Somewhat
   - ○ Moderately
   - ○ Extremely

10. **How did you USE the TOTAL BARS to complete the task?** *

    _____
    _____
    _____
    _____
    _____

11. **Overall, were the words shown RELEVANT?** *

    Mark only one oval.

    - ( ) not at all
    - ( ) slightly
    - ( ) somewhat
    - ( ) moderately
    - ( ) extremely

12. **How CONFIDENT were you with your answers?** *

    Mark only one oval.

    - ( ) not at all
    - ( ) slightly
    - ( ) somewhat
    - ( ) moderately
    - ( ) extremely

13. **Give example(s) when you were LESS CONFIDENT and possibly explain why** *

    _____
    _____
    _____
    _____
    _____

14. **How DIFFICULT did you find the task?** *

    Mark only one oval.

    - ( ) very difficult
    - ( ) difficult
    - ( ) neutral
    - ( ) easy
    - ( ) very easy

15. **Give example(s) of when you found the task DIFFICULT and explain why** *

    _____
    _____
    _____
    _____
    _____

16. **Any other comments**

........................................................................................

........................................................................................

........................................................................................

........................................................................................

........................................................................................

# Post-Questionnaire Results for the 12 participants in C2

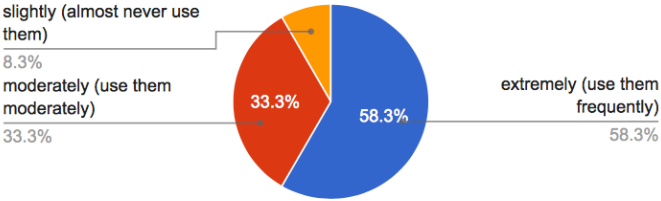

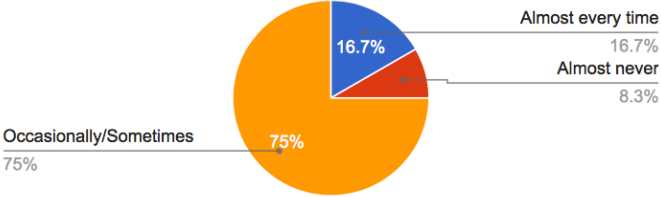

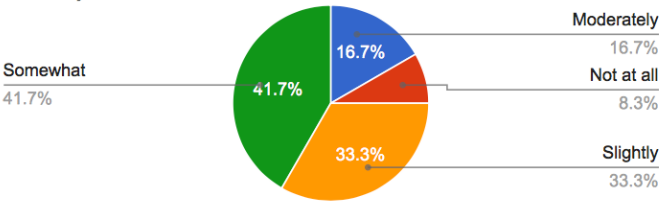

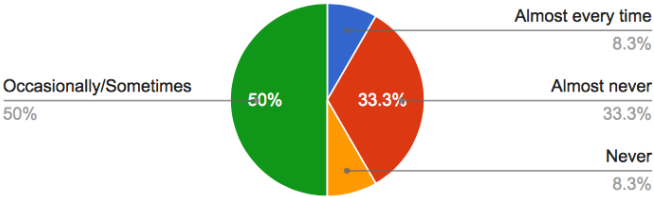

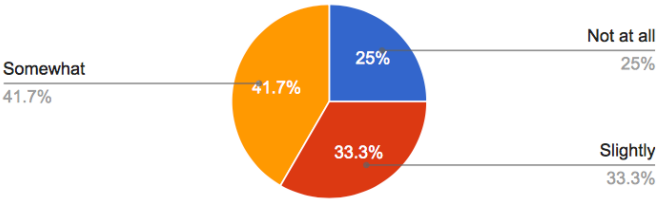

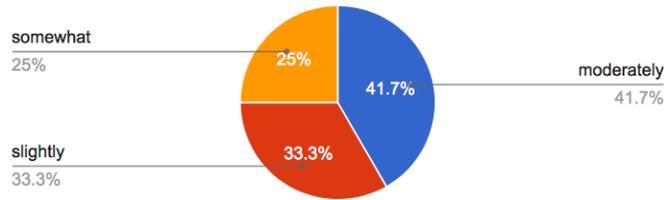

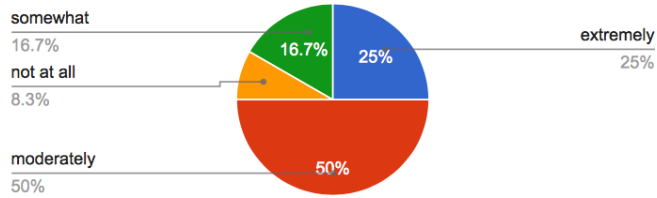

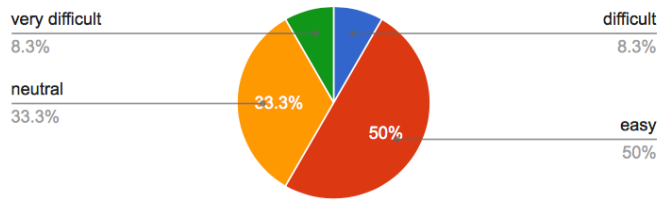

| Which aspects of the User Interface or Visualization (if any) did you find CONFUSING? |
|---|
| First I was not sure about the interpretation of the heat map but it became clear after asking from the experimenter. |
| presented a lot of information that was irrelevant to the task |
| None |
| vertical placement of the words |
| Selection of relevance was confusing as mentioned earlier |
| None |
| None. The visualisations were really good. |
| Clicking the grey bar was somewhat strange in the first time since some times the keyword was quite long and I wondered why would just put some yes-no answers to it. |
| Strange way of choosing the 0 or 1 (no visible button, checkbox), strange way of indicating what I have chosen. Concerning visualisation, the bars (indicating total) were easily confused as indicating relative importance, even though I consciously knew they indicate total. |
| I got no feedback on how anything I did affected the system. Having to think about both heat map value and occurrence number is a bit confusing. |
| None |
| It was not confusing |

| **How did you USE the data shown in HEATMAP to complete the task?** |
|---|
| I checked whether the heat map supported my answer, and if not, I reconsidered and might change my answer. |
| mostly just compared to my selections |
| If the word appeared in enough documents, then checked where it usually appears. It seemed to me though that there were more articles from AI than from the others, because AI seemed to be often with the highest occurrence. |
| I first used my intuition to make the words as relevant or not. Later, for words that were borderline, I checked the heat map to see if my intuition makes sense. |
| If the answer was not immediately clear, I consulted heatmaps. |
| To check if a word that I thought could be used in other fields was or not, how much, and with the count, with how much uncertainty |
| Sometimes I used it cases where I was uncertain. |
| Whenever it was a bit difficult to decide whether that term is also common to other fields than AI, I checked the heat map if that could provide some additional data to be considered additionally. |
| Mostly as a sanity check, to see if my quick subjective judgment had missed something obvious |
| It sometimes affected whether I gave positive or negative feedback, if I was uncertain. |
| When the term is not specific to AI, I check the heatmap. If the value is large, then I mark it as relevant. |
| Few times |

| **How did you USE the TOTAL BARS to complete the task?** |
|---|
| I did not use them basically at all. |
| didn't |
| Yes to check the reliability of the colormap |
| did not use, probably since the totals were too small. |
| If the number of observations was low, I did not trust the results in the heatmaps. |
| Useful in combination with the heatmap to measure uncertainty |
| To assess the reliability of the heat map. |
| I checked them occasionally but did not trust them too much. |
| Not at all |
| Not really. |
| Just once to verify my decision. |
| To see if the words related to AI i.e. learning, Bayes are common across all articles. |

| **Give example(s) when you were LESS CONFIDENT and possibly explain why** |
|---|
| Sometimes it is clear that the word is often used in the field of artificial intelligence, but it might be difficult to know whether or not it appears also frequently in other fields (e.g. test, train, etc.) |
| don't remember |
| I don't remember any exampes |
| There were words like 'recognition' which were borderline and hard to confirm. |
| "Integration" can refer to computing integrals which is common in artificial intelligence but not so much in other domains. However, it can also refer to integrating different (software) systems which is common in domains such as software engineering but not that relevant for artificial intelligence. |
| My knowledge of the other fields is not as strong, so I was unsure if some words I believe to be strongly AI related would also occur often in other fields. |
| Starting from "linear regression" in the introduction: if it is specific to artificial intelligence, it is difficult to draw the border in the future. |
| Sometimes the word clearly is something that is used in AI, such that linear regression but there is really no reason why it would not often occur in other fields too since it is (as well as many other similar words occurring) quite general method. |
| Some words that might have dual meaning, one specific to AI but another that is quite common. |
| I was not that familiar with some of the fields so it was difficult to think if the word was used there or not. |
| "distribution" can mean probability distribution in AI or data distribution in distributed computing |
| For the word "cluster", It can be related to cluster computing or unsupervised learning |

| Give example(s) of when you found the task DIFFICULT and explain why |
|---|
| See the answer to question 'when did you feel less confident'. |
| mostly when just couldn't decide, sometimes not familiar enough with the usage of the term |
| Sometimes needed to think how exclusive the word would be to AI |
| Since the definition of Artificial intelligence is a bit broad, it was confusing, especially since there was the 'data mining' and 'graphics'(or similar) categories. |
| When I was less confident |
| For polysemy or general words I was not sure if they might occur commonly in other fields |
| Almost all of the time. I thought that "artificial intelligence" refers to the subfield of Machine learning and I was looking for words specific to it. This really confused me. The number of papers used for the heat map was constantly so low that I felt obliged to use it much. |
| The same example as in confidence question applies also here. |
| When I was not sure how broadly "AI" is meant in this context... |
| See above. |
| No example |
| This task was easy |

# Post-Questionnaire Results for the 11 participants in C3

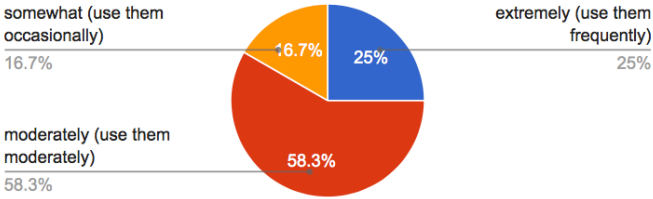

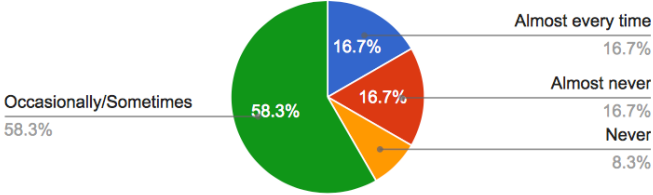

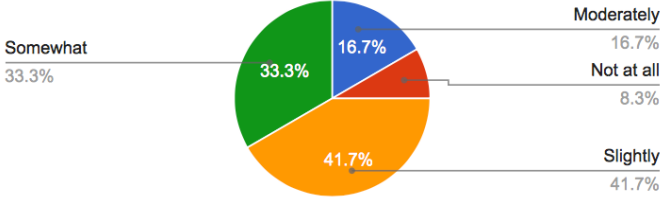

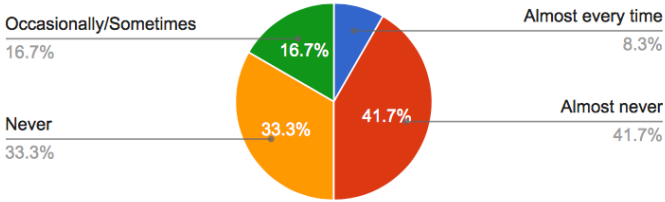

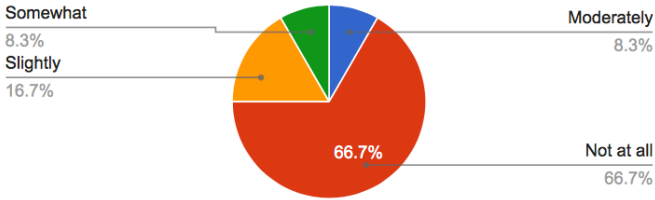

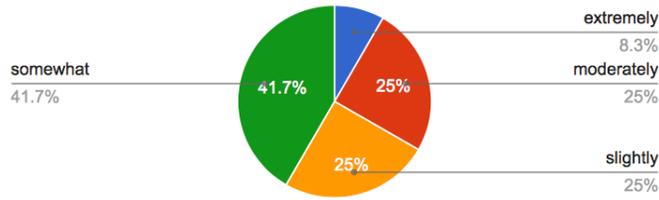

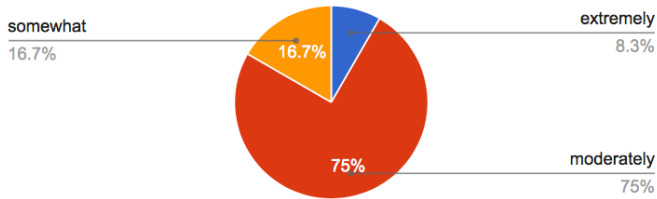

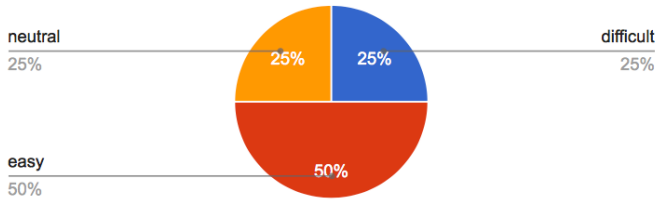

| Which aspects of the User Interface or Visualization (if any) did you find CONFUSING? |
| --- |
| The heat map |
| The total score |
| None, but the text could have been horizontal |
| It was not confusing. |
| small size of components |
| none |
| The second run seems better. |
| I didn't find anything particularly confusing. With respect to completing the given task I didn't make any use of the heatmap and bar chart. In retrospect, I could have made more use of them to guide my decisions. |
| None |
| none |
| I think the visualisation for this task is fine. |

| **How did you USE the data shown in HEATMAP to complete the task?** |
|---|
| When facing an unfamiliar word |
| When I am very doubtful I checked it |
| It affected my decision when I was uncertain of the answer otherwise |
| I used it to check if it complies with my own decision or not. Most of the time it did. |
| in cases that I was not familiar with a term, or I could not decide |
| I have not used them |
| finding the possible relevancy of keywords |
| I didn't. |
| After the first few iterations I found it much more easier to do the task without consulting the heatmap. Initially I was using the heatmap to double check my opinion about the keywords to the data generated opinion shown in the heatmap. |
| to consult about terms, which were ambiguous for me |
| When not sure about the relevance of the keywords to the topics |

| **How did you USE the TOTAL BARS to complete the task?** |
|---|
| I did not use it |
| I rarely used it |
| Did not |
| I used it to see how frequent the word was used. |
| in cases that I was not familiar with a term, or I could not decide |
| I have not used them |
| Just to make sure how reliable the heat maps are that is when the heat map shows a high value, how it matches the total bar value with respect to the total amount of data |
| I didn't. |
| I didn't use the total bars for the task since a more keywords were zero in the total bar. |
| when I looked into the hitmap to check an ambiguous term and hitmap didnt match my expectations |
| I mainly used the heat map when necessary |

| **Give example(s) when you were LESS CONFIDENT and possibly explain why** |
|---|
| The words that were unfamiliar to me |
| There are some words that could also be in the computer graphics domain. I referred to the heat map for such words and then as the colour is similar (but not dark purple= 0.00 relative frequency) for both AI and comp. graphics i decided to score them 0. but had doubt when all the items in the heat map has the same colour (dark purple = 0.00 relative frequency) and when I'm not sure whether it is an AI or comp. graphic related keyword |
| Some words (e.g. graph) that may be used more frequently in some other topics |
| For some words, I was not exactly sure if the word was specifically related to AI but not to other fields. |
| in all cases where the keywords were too general in scope |
| "v" very often appears in the AI literature as vector notation but could as well be used as a symbol in other domains. |
| Some of the other topics have overlap with AI as it is a big area of research |
| Mostly I felt that the words that could be associated with AI could very well also have been associated with e.g. data mining. |
| The keywords in this task which I found hard to indicate as relevant were the ones which could be equally relevant to Information retrieval, data mining or ubiquitous computing. |
| words about embeded systems. I'm not an expert in it, such words may refer to software, networks, AI and other topic, I'm not sure about the border |
| Those keywords I have never used and read from papers. |

| Give example(s) of when you found the task DIFFICULT and explain why |
|---|
| When there were domain overlaps |
| when there are overlapping keywords. |
| Some words (e.g. graph) that may be used more frequently in some other topics |
| This task was easier than the previous one because I had background knowledge myself and I was more confident with my answers with less need to use the heat map etc. |
| contextual ambiguity of words |
| I found the task difficult in the cases when a given word appears often in AI literature but is still not very specific, and could as well be used in other areas, e.g. "evaluation". |
| There was not much difficult situation, there were cases that the term seemed relevant but there were no statistics available for it. |
| See previous answer. The most difficult part was to decide whether a word should be exclusively relevant for AI (in which case almost no words would have been denoted as relevant). That's why some of the answers could have gone either way. |
| I didn't find the task difficult except where the categories of the domain were very similar. |
| ambiguous words, which can be used as very specific terms in different areas |
| Those keywords I have never used and read from papers. |



\end{document}